\documentclass{article}

\usepackage{microtype}
\usepackage{graphicx}
\usepackage{subfigure}
\usepackage{booktabs} 

\usepackage{hyperref}

\usepackage[accepted]{icml2024}

\usepackage{amsmath}
\usepackage{amssymb}
\usepackage{mathtools}
\usepackage{amsthm}

\usepackage[capitalize,noabbrev]{cleveref}

\theoremstyle{plain}

\theoremstyle{definition}

\theoremstyle{remark}

\usepackage[textsize=tiny]{todonotes}

\icmltitlerunning{Reasoning Capacity in Multi-Agent Systems}

\newcommand{\size}[2]{{\fontsize{#1}{0}\selectfont#2}}
\newcommand{\para}[1]{\vspace{0.0cm}\noindent\textbf{#1}}

\usepackage{tcolorbox}
\usepackage{tikz}

\newcommand{\postspace}{\vskip -3mm}
\newcommand{\minipostspace}{\vskip -2mm}

\begin{document}

\twocolumn[
\icmltitle{Reasoning Capacity in Multi-Agent Systems: Limitations, Challenges and Human-Centered Solutions}



\icmlsetsymbol{equal}{*}

\begin{icmlauthorlist}
\icmlauthor{Pouya Pezeshkpour}{yyy}
\icmlauthor{Eser Kandogan}{yyy}
\icmlauthor{Nikita Bhutani}{yyy}
\icmlauthor{Sajjadur Rahman}{yyy}
\\
\icmlauthor{Tom Mitchell}{comp}
\icmlauthor{Estevam Hruschka}{yyy}
\end{icmlauthorlist}

\icmlaffiliation{yyy}{Megagon Labs}
\icmlaffiliation{comp}{Carnegie Mellon University}

\icmlcorrespondingauthor{Pouya Pezeshkpour}{pouya@megagon.ai}

\vskip 0.3in
]



\printAffiliationsAndNotice{}
\begin{abstract}

Remarkable performance of large language models (LLMs) in a variety of tasks brings forth many opportunities as well as challenges of utilizing them in production settings. Towards practical adoption of LLMs, multi-agent systems hold great promise to augment, integrate, and orchestrate LLMs in the larger context of enterprise platforms that use existing proprietary data and models to tackle complex real-world tasks. Despite the tremendous success of these systems, current approaches rely on narrow, single-focus objectives for optimization and evaluation, often overlooking potential constraints in real-world scenarios, including restricted budgets, resources and time. Furthermore, interpreting, analyzing, and debugging these systems requires different components to be evaluated in relation to one another. This demand is currently not feasible with existing methodologies. 
In this postion paper, we introduce the concept of \textit{reasoning capacity} as a unifying criterion to enable integration of constraints during optimization and establish connections among different components within the system, which also enable a more holistic and comprehensive approach to evaluation. We present a formal definition of reasoning capacity and illustrate its utility in identifying limitations within each component of the system. We then argue how these limitations can be addressed with a self-reflective process wherein human-feedback is used to alleviate shortcomings in reasoning and enhance overall consistency of the system. 

\end{abstract}

\section{Introduction}

Recent years have witnessed tremendous advancements in large language models (LLMs), showcasing notable capabilities in addressing a wide array of tasks solely through natural language instructions and minimal in-context examples. These have facilitated development of multi-agent systems, where multiple agents with distinct tasks and models can communicate and collaborate to perform complex tasks \cite{schick2023toolformer, lu2023chameleon, paranjape2023art, liang2023taskmatrix}. The agents within these systems vary from LLMs to domain-specific models, tools and APIs. Such systems are especially appealing for the real-world deployment of LLMs where incorporating proprietary data and models, while considering constraints such as latency, performance, costs, and privacy is necessary.


\begin{figure}[t]
  \centering
  \includegraphics[width=\linewidth]{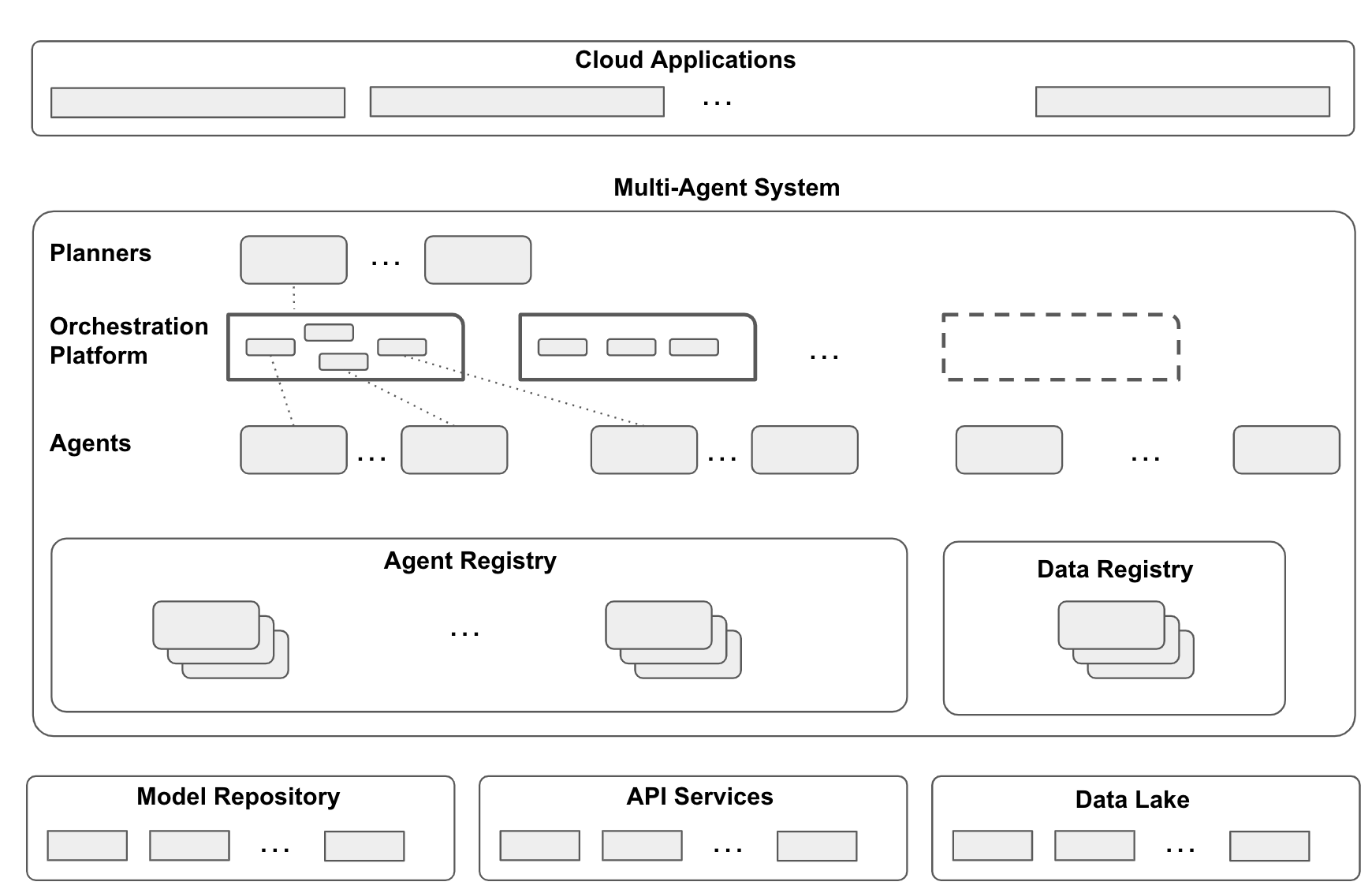}
  \minipostspace
  \caption{Overview of multi-agent systems in an enterprise infrastructure: Cloud applications can utilize multi-agent system to cater for application needs. Multi-agent systems sits in between applications and enterprise infrastructure components such as a model repository, API services, and data lake. Multi-agent systems are comprised of (a) an orchestration platform, (b) planners, and (c) agents. Agent registry keeps record of available agents, while data registry is a repository for data in the enterprise data lake.}
  \label{fig:ow}
\postspace
\minipostspace
\end{figure}

Typically, a multi-agent system consist of three main components: (1) an orchestration platform that facilitates coordination among agents, particularly in terms of data and instruction communication, (2) a planner that generates execution plans by decomposing a complex task into sub-tasks and mapping them to agents, and (3) a set of agents dedicated to solving specific tasks, optionally referring to available data sources. In most recent approaches \cite{schick2023toolformer, lu2023chameleon, hao2023toolkengpt}, a LLM serves as the central component of each module, i.e., a LLM orchestrates plans and invokes agents which  often are LLMs themselves. An example of such systems in an enterprise infrastructure is illustrated in Figure \ref{fig:ow}. 
While these systems exhibit remarkable capabilities, deploying them in production comes with certain challenges. Firstly, relying on a single end-goal objective for optimizing the overall system severely restricts how each component can be optimized, thereby affecting their efficiency and performance. Enforcing production-related constraints, such as efficiency, cost, and privacy, on such single-objective systems is also challenging.
Secondly, the nuances of interactions among multiple components pose a challenge in comprehensively understanding, analyzing, and debugging these systems.

Unlike the typical evaluation methods for machine learning models, assessing a multi-agent system based solely on its overall task performance is inadequate. 
A more rigorous approach involves creating targeted benchmarks to evaluate and troubleshoot each component within the system individually. However, such a solution not only demands significant resources but also assesses components in isolation, posing challenges in debugging multi-agent systems where issues may arise from interactions between different agents.
%
Alternatively, one could employ global and local reward functions from Multi-Agent Reinforcement Learning (MARL) \cite{schneider1999distributed,bagnell2005local, zang2023automatic}. However, MARL lacks effective methods to estimate how each agent's output, as well as the system's collective output, deviates from the optimal solution. As a result, it becomes imperative to introduce a novel criterion to optimize such systems under various constraints and also to provide a more holistic examination of overall performance.


In this position paper, we introduce a novel concept called \textit{reasoning capacity}, 
with the aim of improving optimization, monitoring, debugging, and evaluation in multi-agent systems. We define the reasoning capacity of a multi-agent system as its ability to effectively collect input data, process information, and produce accurate output for a given task and set of constraints, in comparison to the optimal performance of an ideal system. Drawing inspiration from channel capacity in information theory \cite{cover1999elements}, we formally define reasoning capacity as the ratio of the information content in the input and the system regarding the output, to the information content in the input and the \textit{optimal} system regarding the output, under the given constraints. 
%
%
While both reasoning capacity and computational complexity \cite{arora2009computational} assess a system's efficiency and effectiveness, it's crucial to note that reasoning capacity is a broader and more holistic concept, encompassing the overall ability of a system to perform a range of reasoning tasks.

Taking inspiration from distributed computing, where multiple interconnected computers collaborate to perform tasks \citep{van2017distributed}, we offer a comprehensive breakdown of reasoning capacity across various components of the system. Such a breakdown facilitates a more controllable debugging and optimization of the system. Following this, we examine the existing limitations and constraints within different components of a multi-agent system. We discuss their detection and resolution through reasoning capacity (RC) and present potential research directions for the practical applications of multi-agent systems. We argue that integrating human reasoning into various aspects of multi-agent systems can alleviate the bottlenecks in systems' reasoning mechanism. As a result, we posit that self-reflection with human feedback can enhance reasoning capabilities and overall consistency in multi-agent systems.

\section{Multi-Agent Systems}

Multi-agent systems typically comprise of three primary components: a) an orchestration platform, b) planner(s), and c) agents. Figure \ref{fig:ow} shows the overview of such a platform in a real-world scenario. 
The orchestration platform serves as a central hub for managing data flow and interactions among agents. It provides infrastructure to guide the planner, delegate tasks and exchange information, ensuring efficient and cohesive operation of the system.


The planner decomposes complex tasks into sub-tasks and assigns them to agents through the orchestration platform. This process is essential for the effective distribution of data and instructions. Some prior studies use pre-trained LLMs like GPT directly as planners \cite{lu2023chameleon}, while others employ supervised or reinforcement learning approaches for training a planner \cite{schick2023toolformer,liu2023integration}. Multi-agent systems in production settings can extend these works and employ multiple planners, each generating concurrent plans with different target constraints (e.g., cost, latency). Alternatively, they can also operate without a dedicated planner, relying on agents to self-coordinate with guidance from the orchestration platform.





Agents in multi-agent systems can span from general-purpose agents capable of handling open-ended tasks with models like LLMs, to specialized agents trained on proprietary data for specific tasks. This diversity enables them to address a wide array of challenges and tasks effectively. Undoubtedly, agents  vary in their reasoning capacities as well as in their task performance and production-related characteristics, such as response time.
Planner(s) can use such information in creating task plans, or for optimization purposes. However, existing works do not explore optimization under multiple constraints in real-world scenarios.


\subsection{Multi-Agent System in an HR Enterprise}

In this context, we introduce a real-world example of a multi-agent system to gain an understanding of these systems and the associated real-world constraints. Imagine an HR company, equipped with diverse cloud applications for resume uploads, job vacancy postings, job and candidate searches, and applicant tracking. These applications, integrated into the company's infrastructure, gather and process various data types—resumes, job descriptions, and applications—utilizing APIs to interact models and data. A multi-agent system in this company bridges cloud applications and infrastructure by facilitating advanced tasks such as career guidance, candidate matching and resume writing assistance (refer to Figure \ref{fig:ow}). The system needs to handle queries ranging from open-end questions like career support to more structured tasks like candidate matching. It efficiently coordinates with specialized agents, as well as general data query agents. 


Consider a user query, ``\textit{What are the top skills for a senior machine learning engineer?}'' to this system. To respond to this query, let's assume that the system can access various agents, including an LLM-based question-answering agent (LLMQA), a document-retrieval agent linked to a job-description corpus (JDR), a skills extractor agent (SKILLEX), a text-to-SQL agent (TEXT2SQL), and a SQL-query agent (POSTGRES) managing a relational database with historical applications, extracted resumes, and job descriptions (JDs). Additionally, let's assume the system operates within a specific budget to illustrate real-world constraints. We provide a concrete illustration of a possible planner for this example in the Appendix.

Multiple plans can be devised for such a query. The simplest plan involves querying the LLMQA agent and showing its prediction to the user. A more complex plan relies on the TEXT2SQL agent to convert the query into SQL, and then execute the POSTGRES agent. Yet another plan relies on JDR agent to fetch JDs for the specified role and the SKILLEX agent to extract and summarize the skills. Each of these plans vary in cost, limitations and reasoning capacities. Querying LLMQA might be efficient but does not provide a data-driven response. Translating query to SQL depends on the effectiveness of translation, database schema and content. Extracting directly from JDs is most expensive but can be effective especially for long-tailed job titles. Identifying the most optimal plan considering performance and optimization constraints poses a challenging problem that remains unexplored in existing studies. 

\section{Reasoning Capacity}
\label{sec:rc}

To tackle the challenges outlined earlier, we propose a novel concept \textit{reasoning capacity}. 
Drawing inspiration from channel capacity in information theory, which is defined as the maximum mutual information between input and output with respect to the input distribution \cite{cover1999elements}, 
we define reasoning capacity as the system's overall ability to process input effectively and generate output for a given task and set of constraints. Formally, let's consider a multi-agent system denoted as MAS, which executes a plan $PT$ designed to handle a particular task $T$ under constraint $C$. This task is defined by input-output pairs $X_T$ and $Y_T$. Reasoning capacity $RC_T$ is then defined as: 
%
\begin{align}
    RC_T(\text{MAS}, C) = 
    \frac{I(Y_T; \text{MAS}^{PT}, X_T|C)}{max_{\text{MAS}_{i=1,\text{...}, N}}I(Y_T; \text{MAS}^{PT_i}_i, X_T|C)}\nonumber
\end{align}
where $N$ represent the number of potential systems for that task and  $I(Y_T; \text{MAS}^{PT}, X_T|C)$ denotes the mutual information between the output, system, and input given constraint $C$. In other words, $I$ denotes the information about $Y_T$ gained from observing $\text{MAS}^{PT}$ and $X_T$ while considering the constraint. The aim is to normalize reasoning capacity by contrasting the target system with the optimal system, achieved by maximizing the mutual information across all viable systems $\text{MAS}_i$ and their respective plans $PT_i$ for task $T$.

It's important to note that although we are presenting a mathematical formulation for reasoning capacity in the context of a multi-agent system, we are defining it in relation to its components, such as individual agents, without assigning a specific absolute value to the capacity. Even without a precise approximation of reasoning capacity, we argue that the mathematical formulation can provide valuable insights into addressing existing practical limitations due to its component-based breakdown (refer to Section \ref{sec:break}). 
To approximate reasoning capacity, if labeled data that mirrors 
real-world input distributions for each component is available, one can directly estimate the numerator. When labeled data is unavailable, $RC$ can still be indirectly estimated. One approach to approximate reasoning capacity involves providing input and predicted output pairs from each component to a language model like GPT-4 to generate the component description. RC can then be estimated by comparing this description to the actual component description using an entailment score from a natural language inference (NLI)-based model \cite{storks2019recent}. Alternatively, in systems with multiple agents per subtask, RC can be estimated based on the agents' consensus, akin to the method in \cite{platanios2017estimating}. 
Moreover, aiming for a tight upper bound on the denominator (drawing inspiration from methods like the Bayesian Optimum Classifier \cite{mitchell1997machine}), RC can be utilized as a stand-alone metric. 
While reasoning capacity likely correlates with the system's accuracy and confidence in task $T$, its accurate estimation remains a challenge for future research.

\subsection{Leveraging Reasoning Capacity to Connect Components of a Multi-Agent System}
\label{sec:break}
Reasoning capacity serves as a key metric for assessing the impact of individual component on the overall performance. It offers a cohesive framework to evaluate the system's ability to reason, infer, and make informed decisions across its components. Specifically, it includes (1) \textbf{Planner Reasoning Capacity}, reflecting the planner's capability in task decomposition and agent, and (2) \textbf{Agents Reasoning Capacity}, reflecting the combined reasoning prowess of the agents for the task at hand. 
Planner reasoning capacity is dependent on the planner. We discuss it in more details in Section \ref{sec:plan}. 
For agents' reasoning capacity, drawing inspiration from distributed computing \cite{van2017distributed}, we consider decomposing tasks into parallel and sequential sub-tasks and explore how to aggregate reasoning capacity in each scenario. This approach allows us to recursively determine the overall system's RC.

Consider a block $B$, composed of $n$ sub-blocks $B_i$, where each $B_i$ can be an individual agent or a block of multiple agents. We will first analyze the scenario where these agents or blocks interact in a sequential arrangement:

\begin{center}
\begin{tikzpicture}[scale=0.5, every node/.style={scale=0.6}]
            \node at (0,0) [draw,rectangle,thick,inner sep=0,minimum width=7cm,minimum height = 3cm, fill=gray!8!white, label=above:\size{18}{$B$}] (B){};
            \node at (-2.5,0) [draw,rectangle,thick,inner sep=0,minimum size=2.0cm, fill=gray!20!white] (B1) {\size{18}{$B_1$}};
            \node at (2.5,0) [draw,rectangle,thick,inner sep=0,minimum size=2.0cm, fill=gray!20!white] (Bn) {\size{18}{$B_n$}};
            \draw[dashed, line width=1.5mm, gray!80!black] (B1) -- (Bn);
            \draw[line width=1.5mm, gray!80!black] (Bn) -- (5,0) node [right] {\textcolor{black}{\size{20}{$Y_B$}}};
            \draw[line width=1.5mm, gray!80!black] (B1) -- (-5,0) node [left] {\textcolor{black}{\size{20}{$X_B$}}};
\end{tikzpicture}
\end{center}

Then, we can approximate the overall RC of $B$ on the sub-task $T$ by breaking it down as below, where $RC_{T_i}(B_i, C_i)$ is the reasoning capacity of each component in $B$ under constraint $C_i$, and $F_{\text{seq}}$ is an aggregation function:  
\begin{align}
    RC_T(B, C) = \frac{F_{\text{seq}} (RC_{T_1}(B_1, C_1),...,RC_{T_n}(B_n, C_n))}{max_{\hat{B}_{i=1, 2, ..., N}}I(Y_B; \hat{B}^{PT_i}_i, X_B|C)}\nonumber
\end{align}
where $\hat{B}_i$ is any possible block of agents completing sub-task $T$, and $C_i$s are either breakdown of $C$ in constraint such as budget or equal to $C$ in constraints such as privacy. In a similar fashion, we can formulate reasoning capacity of a block $B$ containing $n$ parallel components, as below, using $F_{\text{parallel}}$:

\begin{center}
\begin{tikzpicture}[scale=0.5, every node/.style={scale=0.6}]
            \node at (0,0) [draw,rectangle,thick,inner sep=0,minimum width=4cm,minimum height = 6.5cm, fill=gray!8!white, label=above:\size{18}{$B$}] (B){};
            \node at (0,-2) [draw,rectangle,thick,inner sep=0,minimum size=2.0cm, fill=gray!20!white] (B1) {\size{18}{$B_1$}};
            \node at (0,2) [draw,rectangle,thick,inner sep=0,minimum size=2.0cm, fill=gray!20!white] (Bn) {\size{18}{$B_n$}};
            \draw[dashed, line width=1.5mm, gray!80!black] (B1) -- (Bn);
            \draw[line width=1.5mm, gray!80!black] (B) -- (3,0) node [right] {\textcolor{black}{\size{20}{$Y_B$}}};
            \draw[line width=1.5mm, gray!80!black] (B) -- (-3,0) node [left] {\textcolor{black}{\size{20}{$X_B$}}};
            \draw[line width=1.5mm, gray!80!black] (-2.45,0) -- (-1.9,0);
            \draw[line width=1.5mm, gray!80!black] (-1.9,-0.1) -- (-1.9,2);
            \draw[line width=1.5mm, gray!80!black] (-2.05,2) -- (-1.15,2);
            \draw[line width=1.5mm, gray!80!black] (-1.9,0.1) -- (-1.9,-2);
            \draw[line width=1.5mm, gray!80!black] (-2.05,-2) -- (-1.15,-2);
            \draw[line width=1.5mm, gray!80!black] (2.45,0) -- (1.9,0);
            \draw[line width=1.5mm, gray!80!black] (1.9,-0.1) -- (1.9,2);
            \draw[line width=1.5mm, gray!80!black] (2.05,2) -- (1.15,2);
            \draw[line width=1.5mm, gray!80!black] (1.9,0.1) -- (1.9,-2);
            \draw[line width=1.5mm, gray!80!black] (2.05,-2) -- (1.15,-2);
\end{tikzpicture}
\end{center}
\begin{align}
    RC_T(B, C) = \frac{F_{\text{parallel}} (RC_{T_1}(B_1, C_1),...,RC_{T_n}(B_n, C_n))}{max_{\hat{B}_{i=1, 2, ..., N}}I(Y_B; \hat{B}^{PT_i}_i, X_B|C)}\nonumber
\end{align}

Let us note that in sequential blocks, the reasoning capacity of earlier agents influences subsequent ones, which should be taken into consideration when approximating $F_{\text{seq}}$. Conversely, in parallel blocks, this inter-agent effect is typically absent.
To approximate aggregation functions, a common approach is to employ multiplication and summation for sequential and parallel blocks, respectively, assuming equal contributions from each component. 
Although suitable when optimizing reasoning capacity as the objective, these approximations may not be optimal for identifying system bottlenecks. Selector functions, such as maximum or minimum, or those choosing a subset of agents, prove more effective for bottleneck detection. Additionally, interpretability methods like Shapley values \cite{lundberg2017unified} or neural network-based approximations can be used to more accurately estimate agents' contributions. However, we leave exploration of these aggregation functions for future research.

\subsection{Reasoning Capacity in the HR Company Example}

For the HR company scenario, planners optimized for accuracy, might overlook budget constraints and prioritize the most probable successful plan. Additionally, if the system selects, for example, the third plan and the SKILLEX agent lacks adequate reasoning capacity to precisely extract skills, the system could struggle to generate accurate responses. Relying solely on overall system accuracy in this situation makes it difficult to pinpoint the underlying problem.


To address these limitations, we will simplify the system's reasoning capacity breakdown. Assuming the planner opts for the third plan involving the JDR and SKILLEX agents sequentially, and we use multiplication for aggregating their capacities. We also approximate each agent's reasoning capacity with an NLI-based model entailment score, applying it to descriptions generated by GPT-4 for each agent's input-output pairs. 
Thus, we can approximate $RC_{T_{\text{JDR}}}(\text{JDR}, \text{budget})$ and $RC_{T_{\text{SKILLEX}}}(\text{SKILLEX}, \text{budget})$ using above assumptions. Then, the overall reasoning capacity of the system for this task can be calculated as follows:
\begin{align}
    \nonumber RC_T(\text{MAS}, b) = 
    \frac{RC_{T_1}(\text{JDR}, b) \times RC_{T_2}(\text{SKILLEX}, b)}{max_{\hat{\text{MAS}}_{i=1,\text{...}, N}}I(Y; \hat{\text{MAS}}^{PT_i}_i, X|b)}\nonumber
\end{align}
where $b$ is the budget constraint, $T$ is the overall task, $T_1$ and $T_2$ are respective tasks for the agents, and $\text{MAS}$ is the target system. 
To better understand the budget constraint's impact on reasoning capacity, let's examine the approximation process of reasoning capacity. We begin by collecting input-output pairs for each component/agent, starting with the system's input. If some instances surpass the budget, they are omitted, reducing the pair count per component. This reduction leads to a less accurate component description and lower entailment scores, effectively decreasing the component's reasoning capacity. Complex components, often causing budget overruns, need more input-output pairs for higher entailment scores. Therefore, identifying the components with the lowest reasoning capacity can pinpoint those responsible for budget breaches. Now that we conceptualize reasoning capacity, in the remainder of paper, we focus on investigating underexplored real-world  limitations in current settings that could potentially be addressed or detected through the utilization of reasoning capacity. 

\section{Bottlenecks in Orchestration Platform}
The orchestration platform, central to coordinating agents and managing data and instructions, is crucial for system performance, especially influencing planner reasoning abilities. It is instrumental in generating viable responses within production constraints, monitoring real-time and historical performance metrics like cost, quality, and response time for each system component. The planner utilizes this data to enforce constraints, including selecting or pruning plans, abandoning, or modifying plans during execution if performance data indicates constraint violations. 
Below, we explore limitations and open problems associated with the orchestration platform, where reasoning capacity can help manage and/or address: 

\para{Dynamic Environment:}
A key challenge for the orchestration platform is maintaining reasoning consistency amidst temporal and external changes in dynamic environments \cite{kim2020multi, liu2023integration, faroni2023optimal}. These changes can originate from (1) compute/storage platforms, (2) data and model artifacts, and (3) external factors. The orchestration platform relies on compute or storage platforms, like cloud services, where changes can significantly affect overall system performance. It is crucial for the orchestration platform to continuously and historically gather data to aid the planner in creating and adjusting plans in response to these changes.

Orchestration platform also facilitates delivery of data and models to agents. Updates to data sources and changes in models versions, can unexpectedly impede the reasoning capabilities of the overall system \cite{chen2023chatgpt}. Reasoning capacity allows us to monitor and assess the effects of these changes on the system, a task that becomes quite challenging if we solely depend on the overall system performance.  
Lastly, external factors such as adversarial attacks \cite{greshake2023not} and human feedback/intervention \cite{paranjape2023art} can lead to a plunge or surge on reasoning capability of the system, making it hard to evaluate/debug the behavior of the system if we only rely on the overall performance. 

\para{Budget:}
Operating under a limited budget is a common constraint for any real-world system. This constraint is particularly crucial when dealing with LLMs through APIs like ChatGPT \cite{chen2023chatgpt} and GPT-4 \cite{openai2023gpt-4}. However, there is always a trade-off that must be made in regards to cost, performance, and quality, when choosing among alternative agents, and data sources.
By approximating the reasoning capacity of different potential agents while considering possible plans, reasoning capacity provides us with the capability to optimize within budget constraints. 
This involves selecting agents with adequate reasoning capacity for the task while also being cost-effective, and meeting overall productions constraints. 
Reasoning capacity can also be used to pinpoint agents causing budget overflows and aid in choosing cost-effective agents with adequate reasoning capacity for a given task during inference.

\para{Ethical/Privacy/Trust Considerations:}
Beyond the discussed limitations, the orchestration platform must adhere to ethics, privacy, and trust considerations. While each agent can target avoiding social biases and harmful content \cite{blodgett2020language}, and privacy requirements, doing so collectively, when agents share data and instructions is yet another challenge, encompassing the proper handling of sensitive information and data protection \cite{mireshghallah2020privacy}, among trusted agents. 
Addressing these challenges necessitates a system-wide approach, as enhancing individual agents might be inadequate. Reasoning capacity is instrumental in both optimizing the system within these constraints and pinpointing sub-optimal components in these areas by identifying bottlenecks during inference. 

\section{Bottlenecks in Planning}
\label{sec:plan}
The planner serves as the backbone of any multi-agent system, and its capabilities significantly influence the reasoning capacity of a system. 
While previously introduced planners \cite{schick2023toolformer,liu2023integration,lu2023chameleon} have demonstrated successful performance, the intricate nature of multi-agent systems makes it challenging to discern issues originating from the planner versus those for which agents are responsible. 
In this section, our goal is to explore the applicability of reasoning capacity to be used as a diagnostic tool in identifying common issues/limitations caused by the planner in multi-agent systems.

\para{Planner Limited Capabilities:}
A key constraint in reaching the system's desired reasoning capacity originates from the planner's limited reasoning power, crucial for task decomposition and coordination. 
Detecting this limitation can be challenging, as the system may perform adequately in simpler tasks but struggle in more complex scenarios. Hence, relying solely on the overall system performance makes it exceedingly challenging to distinguish between issues stemming from agents' capabilities and those arising from planner failures.
In such scenarios, reasoning capacity can serve as a diagnostic tool to identify planner limitations. If the planner ineffectively allocates tasks, the overall performance may decline, even though the components' reasoning capacities might still be high. This disparity between performance and capacity can indicate shortcomings in the planner's capability in task allocation.

\para{Limited Available Agents:}
In scenarios where agent availability is constrained, essential planner functions like task distribution and agent coordination may be impeded. This can lead to suboptimal task allocation and backlogs, causing delays in task completion. Reasoning capacity becomes crucial in such situations, helping identify bottleneck agents in executed plans and guiding the integration of more proficient agents to replace less adequate ones. Additionally, it can play a role in the autonomous and automatic creation of new agents and tasks \cite{mitchell2018never}.



\para{Planner Limited Domain Specific Knowledge:}
Optimal workload distribution by the planner relies heavily on substantial domain-specific knowledge. In real-world scenarios where planners lack comprehensive domain-specific knowledge due to under-representation or limited resources, they may struggle to make informed decisions about task distribution. This knowledge shortfall can lead to improper task assignments to agents, resulting in a decline in overall system performance. Unlike issues arising from the limited capabilities of the planner, in this scenario, the reasoning capacity of various components in the system will consistently exhibit low performance due to the planner assigning incompatible tasks to agents. This synchronized low performance across the overall system and the reasoning capacity of different components can help identify gaps in the planner's domain-specific knowledge.

\section{Bottlenecks in Intelligent Agents (LLMs)}

This section explores underexplored limitations of agents in multi-agent systems, with a particular focus on Large Language Models (LLMs) due to their essential role in reasoning and problem-solving. LLM agents are known for their proficiency in understanding and generating text, rendering them valuable components in various applications. We describe potential limitations of LLM agents in existing settings and discuss how reasoning capacity can serve as a solution to these limitations.


\para{Out of Distribution Task/Data:}
When incorporating LLMs into multi-agent systems, challenges may arise from out-of-distribution tasks or data. Despite excelling in natural language processing, LLMs may struggle with content beyond their trained domains \cite{bang2023multitask}, leading to difficulties in handling novel or infrequent scenarios. This limitation can result in LLMs generating inaccurate or irrelevant responses, causing reasoning inconsistencies within the system.

To demonstrate this limitation, consider a scenario with two individuals who share a name: one famous, one not. LLMs, primarily trained on data about the famous individual, might provide inaccurate responses to queries about the lesser-known person by referencing the well-known one. For instance, ``Calvin Russell'' refers to both a renowned musician from Austin, Texas, and a less-known football player from Fairburn, Georgia. If we inquire GPT-4 about the birthplace of the football player, it responds with:
\begin{tcolorbox}[fontupper=\ttfamily]
    \scriptsize
    \# prompt:\\
    Where was Calvin Russell born in?\\
    \# GPT-4 Output:\\
    Calvin Russell was born in Austin, Texas, USA.
\end{tcolorbox}

\para{Self-Verification Capability:}
Another limitation affecting LLMs-based agents is the absence of self-verification capability, particularly crucial in tasks involving sensitive information. Despite their proficiency in generating content, LLMs often lack the inherent ability to validate the accuracy or credibility of the information they produce. This limitation can be especially problematic when dealing with tasks where data integrity and reliability are paramount, as the system may generate content without the means to verify its correctness. Although previous works have explored ways to improve self-verification capability \cite{mitchell2018never,sodhani2020toward,elazar2021measuring}, these solutions have not yet reached a consolidated state and most of them are solely based on measuring the self-consistency of the system.


As an illustration, let's examine a disease-symptom correlation task where we provide GPT-4 with a list of symptoms and request predictions for the two most likely diseases. Consider the following example:
\begin{tcolorbox}[fontupper=\ttfamily]
    \scriptsize
    \# prompt:\\
    what are top 2 possible diseases associated with the symptoms of `back pain, weakness in limbs, neck pain, dizziness, loss of balance'. List possible diseases without extra explanation.\\
    \# GPT-4 Output:\\
    1. Multiple Sclerosis, 2. Cervical Spondylosis
\end{tcolorbox}

The provided symptoms align more with "Cervical Spondylosis," the second option, demonstrating the necessity of a self-verification mechanism to ensure the trustworthiness of predictions made by LLMs.


\para{Disputing Agents:}
LLM-based agents can also have conflicts. These may arise from various reasons, including outdated or unreliable sources, instances of hallucination in LLM-generated content, or the inherent imperfections in the agents. Such disputes can lead to inconsistencies in reasoning within the system, as different agents may offer conflicting information or interpretations. An example of this limitation is observed in \citet{mallen2022not}, where the authors demonstrate that LLMs perform better in answering questions about popular entities without being augmented with additional pieces of evidence. Thus, if we inquire about a popular entity from the system, and the planner includes both the LLM and the augmented-LLM agents in its plan, conflicts in the output may arise. In such cases, the system needs a confidence measurement to resolve these conflicts.


\para{Tyrant Planner:} 
The reasoning capability of LLM agents might also be compressed as a result of a tyrant planner. This scenario may surface when a planner, having access to numerous available agents, overfits to complex tasks. However, when confronted with simpler tasks, the planner's confidence in its decisions may fall short of the collective confidence of one or a combination of agents. In such cases, these agents should ideally have the capability to overthrow or outvote the planner's final output, ensuring that the reasoning capacity aligns with the most reliable and accurate source. 


To illustrate both the limitations concerning disputes and the potential influence of a tyrant planner, consider the handling of temporal information that evolves over time. In this scenario, we assume that when external knowledge is accessible, the planner might favor augmenting LLMs rather than solely relying on the information stored within LLMs to answer a question. Consider the question to GPT-4: 
``which award did X win most recently?'' compared to an augmented query: ``X won the award Y. Which award did X win most recently?''.  GPT-4 responses are as follows: 


\begin{tcolorbox}[fontupper=\ttfamily]
    \scriptsize
    \textbf{LLM Agent}\\
    \# prompt:\\
    which prize did James A. Michener win most recently?\\
    \# GPT-4 Output:\\
    James A. Michener passed away in 1997, so he hasn't won any recent awards. However, the last major award he received while alive was the Presidential Medal of Freedom in 1977.\\
    \\
    -------------------------------------------------\\
    \textbf{Augmented LLM Agent}\\
    \# prompt:\\
    James A. Michener won the prize Pulitzer Prize for Fiction. Which prize did James A. Michener win most recently?\\
    \# GPT-4 Output:\\
    James A. Michener's most recent major award was the Pulitzer Prize for Fiction in 1948 for his book "Tales of the South Pacific". He passed away in 1997 and did not receive any other major literary awards after the Pulitzer.
\end{tcolorbox}
The accurate response to this question is ``Presidential Medal of Freedom,'' which GPT-4 was able to capture without any external augmentation,  showcasing the necessity of a confidence mechanism to choose the LLM agent's response over the augmented LLM agent.

To address these limitations, it is necessary to provide a reliable confidence measurement for any given instance, one that is applicable to every component/agent within the system. However, reasoning capacity is inherently designed to capture the overall behavior of a system/component rather than instance-level behaviors. To integrate reasoning capacity and approximate the confidence of a single data point for any system component, we need to estimate the instance-level reasoning capacity of the component for that specific data point.
If acquiring instance-level values from the approximation of reasoning capacity becomes challenging without needing knowledge of the gold output, one viable approach is to employ a K-NN \cite{guo2003knn}-based approximation, a method akin to previous works that adopted a similar methodology to approximate entropy \cite{sricharan2011k}. This involves estimating the reasoning capacity of the target point by relying on the available reasoning capacity of its k-nearest neighbors. Once the reasoning capacity for the target sample is approximated, it can be utilized to identify out-of-distribution cases, self-verify responses, resolve disputes between agents, and overthrow a tyrant planner.


\section{Self-Reflecting with Human Feedback}
Having established the utility of reasoning capacity in identifying limitations within multi-agent systems, we propose the adoption of a self-reflective mechanism to resolve those limitations. More specifically, when the system encounters bottlenecks in the reasoning capacity of different components, it can reflect on those limitations by initiating the use of human reasoning capabilities to resolve them.  
Humans can enhance the system by (1) \textbf{Adding new agents/data} to improve system's capabilities, (2) \textbf{Directly assisting with a task as an agent} by providing complete or partial solutions for a task, and (3) \textbf{Offering feedback} to refine the system, addressing specific issues with instances, agents, or the planner. Note that we assume the humans operating within the multi-agent setting are experts, selected via well-studied approaches in HCI literature such as ability discovery~\cite{chen2023hitsndiffs}.

\para{Constraints Imposed on Orchestration Platform:} 
Suppose a breach occurs in one of the constraints, such as budgetary, ethical, or privacy considerations imposed on the system. We have previously discussed the capability of reasoning capacity in identifying the responsible components or agents in such scenarios.   
After identifying the culprit, there are various ways to involve humans in resolving the issue. One option is to replace the problematic component with a human agent, akin to customer service. While a human with appropriate training can adhere to most imposed constraints while solving the task, this solution is not cost- and time-effective, specifically in a highly user-interactive production setting, and is likely only applicable in very sensitive or complex scenarios. 
Alternatively, an expert human can suggest new agents satisfying imposed constraints to replace problematic agents or provide high-level feedback to align the agent \cite{wang2023aligning} to adhere to the constraints.

\para{Mitigating Planner Limitations:}
In response to limitations imposed by the planner's inadequacy, integrating human reasoning can provide an invaluable solution. Human reasoning can be leveraged to validate and augment the planner's decisions, especially in scenarios that demand a deep understanding of the task's context. Humans can directly assist the planner not only in decomposing complex tasks into manageable sub-tasks and assigning them to available agents but can also introduce new agents if necessary. 
In contrast to instance-level issues, where utilizing humans as agents can be very expensive, high-level human interaction with the planner in the form of an advisory agent can be cost-effective. 
Similarly, humans can provide high-level feedback as training data to further fine-tune the planner, resolving potential limitations.  

\para{Agents Limitations:}
Human feedback can play a pivotal role in addressing agents' reasoning limitations, particularly for LLMs. It can offer validation and fact-checking, crucial for maintaining data integrity in tasks where accuracy is crucial. In out-of-distribution cases, humans can introduce new agents or provide annotations to adapt existing ones to new domains \cite{gururangan2020don}. Additionally, human input is vital in mediating agent disputes, acting as a judge agent to evaluate information credibility and relevance, guiding the system towards informed consensus, and resolving conflicts effectively.

Integrating human reasoning into multi-agent systems proves effective in mitigating their inherent limitations. With the significant role of human feedback in addressing reasoning-related challenges, enhancing human-in-the-loop mechanisms becomes crucial. Recent advancements in Reinforcement Learning with Human Feedback (RLHF) \cite{christiano2017deep, ouyang2022training} offer promising methodologies for aligning LLMs with human preferences, both stylistically and ethically \cite{bai2022training, ganguli2022red}. By harnessing LLMs' extensive capabilities, RLHF fosters responses and behaviors that resonate with human values, contributing to the development of safer, more effective, and controllable AI systems.

Existing methods only collect coarse-grained feedback such as binary acceptability of an AI systems output. In the context of reasoning capacity, richer human feedback on such constraints are required to effectively fine-tune the planner and agents. Moreover, not only we require to harness human expertise to enhance individual agents' reasoning but also foster a collaborative environment that leverages human judgment, validation, and conflict resolution, ensuring the system's consistent and reliable performance across diverse domains and challenges. However, to enable seamless integration of such human-centered design within a production setting would require practitioners to revisit and instrument classical distributed system principles~\cite{fox1999harvest,gilbert2012perspectives,gilbert2002brewer}.
By bridging the gap between computational and human intelligence, we can empowers multi-agent systems to operate with greater adaptability and robustness, thus resolving the limitations identified using reasoning capacity. 

\section{Related Work}
Current endeavors in multi-agent systems predominantly concentrate on enhancing the accuracy and efficiency of planners. Previous works typically revolve around two planner design approaches: (1) training a model, often a large language model, in a supervised manner with labels representing execution plans for various queries, or employing reinforcement learning techniques to train the planner \cite{schick2023toolformer,liu2023integration}. (2) Employing an off-the-shelf large language model, such as GPT-4, as the planner, tasked with generating execution plans in a zero- or few-shot setting \cite{lu2023chameleon}. To further enhance these planners, prior works also explore autonomous self-reflective mechanisms \cite{shinn2023reflexion, li2023zero}. 
Let us note, that the multi-agent framework explored in this study differs from bio-inspired multi-agent systems \citep{leitao2012bio} such as Swarm Intelligence~\citep{kennedy2006swarm}, where a multitude of agents interact to collectively demonstrate intelligent behavior, and Self-Organization Mechanisms \citep{ye2016survey}, where agents autonomously organize and coordinate their actions to achieve collective goals without central control. In contrast, the multi-agent systems discussed in this work revolve around the design and deployment of multiple agents collaborating toward a common objective under central guidance.

\section{Conclusion}
We introduce a novel concept of \textit{reasoning capacity}, positioning it as a pivotal metric for the successful implementation of multi-agent systems in practical scenarios. Through a comprehensive exploration of reasoning capacity and its breakdown across system components, we demonstrate its efficacy in identifying and addressing limitations within multi-agent systems. Reasoning capacity not only can help with enhancing system robustness and efficacy but also can pave the way for more sophisticated solutions in complex, dynamic environments. Moreover, we advocate for a self-reflective process that incorporates human feedback as a promising strategy to compensate for reasoning shortcomings and elevate overall system consistency. This work lays the foundation for a more holistic and human-centered approach to address the challenges and limitations inherent in complex, real-world settings.

\section{Broader Impact} 
The proposed framework in this work has far-reaching implications not only for the technical development of multi-agent systems but also for their ethical and societal impact. By introducing reasoning capacity as a unifying criterion, we lay the groundwork for addressing real-world constraints and limitations that are often overlooked in the development process. This approach can contribute to the responsible and ethical deployment of multi-agent systems, ensuring that they operate within the privacy considerations, and ethical guidelines. Additionally, the integration of human reasoning into the self-reflective process offers a unique opportunity to enhance these systems' ethical and moral decision-making capabilities. As multi-agent systems continue to play an increasingly significant role in various domains, this framework sets the stage for a more responsible and socially conscious approach to their development, leading to systems that align with human values and societal needs.


\bibliography{main}
\bibliographystyle{icml2024}

\newpage
\appendix
\section{Planer Illustration}
To offer a concrete illustration of the planner, we adopt the proposed approach from \cite{lu2023chameleon}. In the context of the provided example, we can define our planner as GPT-4 tasked with providing a sequential plan for executing agents, given the description of agents and the query. We can have GPT-4 conducting such planning in a zero-shot setting, providing an example prompt such as:

\begin{tcolorbox}[fontupper=\ttfamily]
    \scriptsize
    \# prompt:\\
    You need to act as a policy model, that given a question and a modular set, determines the sequence of modules that can be executed sequentially to solve the question.\\
    \\
    The modules are defined as follows:\\
    \textbf{LLMQA:} This module generates an answer for a given question using a pre-trained large language model. \\
    \textbf{JDR:} This module retrieve a relevant document from a job-description corpus given a context.\\
    \textbf{SKILLEX:} This module extract list of skills provided in a given document.\\
    \textbf{TEXT2SQL:} This module convert human-readable text queries into SQL queries.\\
    \textbf{POSTGRES:} This module retrieves responses for SQL queries from a relational database that includes historical applications, along with extracted resumes and job descriptions.\\
\\
    \textbf{Question:} What are the top skills for a senior machine learning engineer?
\end{tcolorbox}


\end{document}